\documentclass{article}

\usepackage[preprint]{neurips_2026}

\usepackage[utf8]{inputenc}
\usepackage[T1]{fontenc}
\usepackage[hidelinks]{hyperref}
\usepackage{url}
\usepackage{svg}
\usepackage{booktabs}
\usepackage{amsfonts}
\usepackage{amsmath}
\usepackage{amssymb}
\usepackage{nicefrac}
\usepackage{microtype}
\usepackage{xcolor}
\usepackage{graphicx}
\usepackage{enumitem}
\usepackage{algorithm}
\usepackage{algpseudocode}
\usepackage{amsthm}
\usepackage{bm}
\usepackage{bbm}
\usepackage{multirow}
\usepackage{longtable}
\newtheorem{pro position}{Proposition}
\graphicspath{{images/}}
\svgpath{{figures/}{images/}}
\usepackage{wrapfig}
\title{TSFMAudit: Data Contamination Auditing in Forecasting Time Series Foundation Models}

\author{%
  Hongkai Li\thanks{Equal contribution.}\\
  Zhejiang University\\
  \texttt{22521276@zju.edu.cn}
  \And
  Shifeng Xie\footnotemark[1]\\
  T\'el\'ecom Paris\\
  \texttt{shifeng.xie@telecom-paris.fr}
  \And
  Lefei Shen\\
  Zhejiang University\\
  \texttt{lefeishen@zju.edu.cn}
  \And
  Zhuo Li\\
  State Street Technology (Zhejiang) Ltd.\\
  \texttt{lizhuo@zju.edu.cn}
  \And
  Mouxiang Chen\\
  Zhejiang University\\
  \texttt{chenmx@zju.edu.cn}
  \And
  Xiaobin Zhang\\
  Zhejiang University\\
  \texttt{22421162@zju.edu.cn}
  \And
  Han Fu\\
  Zhejiang University\\
  \texttt{11821003@zju.edu.cn}
  \And
  Jianling Sun\\
  Zhejiang University\\
  \texttt{sunjl@zju.edu.cn}
  \And
  Xiaoxue Ren\\
  Zhejiang University\\
  \texttt{xxren@zju.edu.cn}
  \And
  Chenghao Liu\thanks{Corresponding author. This work was conducted before Chenghao Liu joined Datadog.}\\
  Datadog\\
  \texttt{twinsken@gmail.com}
}

\begin{document}

\maketitle

\begin{abstract}
Time series foundation models (TSFMs) are increasingly pretrained on large corpora, raising concerns that evaluation datasets may have been exposed during pretraining and thus yield overly optimistic performance estimates. Auditing such contamination is challenging in time series because signals are continuous and heterogeneous, and often lack corpus documentation. To the best of our knowledge, this is the first work to study pretraining contamination auditing for TSFMs. We formalize the problem of pretraining contamination auditing for TSFMs and propose \textsc{TSFMAudit}, a method based on probe adaptation dynamics. Our key intuition is that contamination manifests as unusually efficient adaptation: after a fine tuning probe, contaminated datasets tend to exhibit faster loss reduction with smaller backbone movement. 
We evaluate \textsc{TSFMAudit} on 6 TSFMs and 187 datasets using documented training source evidence as supervision, and compare against 10 competitive baselines adapted from the LLM literature.
\end{abstract}

\section{Introduction}

Time series foundation models (TSFMs) have recently attracted growing attention because of their strong cross domain capabilities across applications such as finance, transportation, and health\citep{li2025tsfm,li2026mira,rahimikia2025re,pulido2026time}.
Although there have been promising attempts to train TSFMs on purely synthetic data, current state-of-the-art models still rely heavily on large scale real world corpora collected from diverse sources\citep{goswami2024moment,xie2026cauker}. At the same time, widely used evaluation benchmarks, are themselves assembled from multiple public sources\citep{aksu2024gift,qiao2026time}. This shared multi-source pretraining evaluation pipeline creates a substantial risk of contamination: data used for evaluation may already have been exposed during pretraining\citep{meyer2025rethinking}.

Such exposure can artificially inflate downstream performance
, and thereby bias our assessment of TSFM generalization\citep{meyer2025rethinking,aksu2024gift,dong2024generalization}. More importantly, contamination may arise unintentionally\citep{meyer2025rethinking,xu2024benchmark}. In time series data, the same underlying signal can reappear under rescaling, renaming, rewindowing, or other preprocessing changes while preserving nearly identical content\citep{meyer2025rethinking,godahewa2021monash}. For example, the Monash archive's Elecdemand dataset is a $1/1000$-scaled subset of Australian Electricity Demand\citep{meyer2025rethinking}. To ensure the fairness and validity of TSFM benchmarking, we therefore need efficient methods for contamination auditing\citep{meyer2025rethinking,aksu2024gift,dekoninck2024constat,kds2025}.


Contamination auditing has been studied extensively in large language models (LLMs) and related domains. Popular families of methods include overlap detection, which searches for direct or near-duplicate benchmark content, memorization extraction, which tests whether a model can reproduce training examples, perplexity-based tests, which flag unusually low model loss, and membership inference attacks, which calibrate a score for deciding whether an example was used in training, often using shadow models, i.e., auxiliary models trained on controlled data splits\citep{yang2023rephrased,shi2024detecting,carlini2021extracting,shokri2017membership,carlini2022lira}. However, these methods do not transfer naturally to TSFMs because time series data are continuous, heterogeneous, and often only partially documented at the corpus level\citep{woo2024moirai,das2024timesfm,qiu2024tfb,garza2024timegpt,goswami2024moment}. For instance, leakage can indeed lead to abnormally low loss, and methods such as \citet{dekoninck2024constat} exploit this signal for contamination detection; however, a smooth seasonal series may also yield loss far below the normal generalization level even when the model has never seen it before, making static loss or perplexity-style criteria unreliable\citep{liu2022losstrajectory,jiang2024investigating}. Likewise, membership inference attacks typically require multiple shadow models to calibrate membership scores, but reproducing TSFM pretraining at that scale is expensive\citep{shokri2017membership,carlini2022lira,hayes2025strongmia}. These limitations call for a contamination auditing method designed specifically for TSFMs.

We make the following empirical observation: under a fine tuning probe, contaminated datasets tend to exhibit faster loss reduction and require less backbone movement than clean datasets as shown at Fig.~\ref{fig:trace} and confirmed by the feature ablation in Table~\ref{tab:feature_ablation}. 
Motivated by this phenomenon, we propose \textsc{TSFMAudit}, a contamination auditing framework for TSFMs based on learning dynamics. \textsc{TSFMAudit} extracts probe time signals such as loss reduction and parameter displacement to infer contamination risk. We summarize our contributions as follows:
\begin{itemize}
    \item To the best of our knowledge, this is the first work to study pretraining contamination auditing for TSFMs. We formalize the dataset level audit objective for contamination.
    \item We propose \textsc{TSFMAudit}\footnote{Code is available at: \url{https://github.com/kkevin117/TSFMAudit}.}, a contamination auditing framework specifically designed for TSFMs, which infers contamination risk from probe time adaptation dynamics.
    \item We evaluate \textsc{TSFMAudit} on 6 TSFMs using their reported pretraining corpora, GIFT-Eval \citep{aksu2024gift}, and the TIME \citep{qiao2026time} benchmark, covering 187 datasets in total. Compared with 10 contamination auditing baselines adapted from the LLM literature, \textsc{TSFMAudit} achieves consistently stronger detection performance.
\end{itemize}

\section{Related Work}
\paragraph{TSFMs and Benchmarks Contamination Concerns.}
Numerous recent works on time series forecasting have moved from small, dataset-specific evaluations toward standardized large-scale benchmarks such as M4, LSF, and the Monash archive, and, with the rise of time series foundation models, further toward broad zero-shot evaluation suites such as GIFT-Eval, TSFM-Bench, and TIME \cite{makridakis2020m4,godahewa2021monash,wu2021autoformer,aksu2024gift,li2025tsfm,qiao2026time}. In parallel, TSFMs themselves are increasingly trained on large heterogeneous corpora collected from diverse real-world and synthetic sources, reflecting the field's ambition to build universal forecasters across domains \cite{ansari2024chronos,das2024timesfm,woo2024moirai,auer2025tirex}. However, this development has also raised a new concern about benchmark validity: many modern benchmarks are assembled by reusing public datasets, evaluation data may overlap with, or remain implicitly connected to, pretraining corpora, leading to hidden information leakage and overly optimistic estimates of zero-shot generalization \cite{aksu2024gift,meyer2025rethinking}. This problem is particularly acute in time series, where fragmented naming and versioning, weak semantic transparency, and strong temporal dependence make contamination harder to detect than in other modalities \cite{meyer2025rethinking}. These concerns have recently motivated a new wave of benchmark design centered on fresher data sources and stricter evaluation protocols intended to reduce leakage and restore the credibility of TSFM comparison \cite{aksu2024gift,qiao2026time}.

\paragraph{Contamination Auditing in LLMs.}
Large language models have motivated a broad literature on contamination auditing. Existing methods mainly fall into four families: corpus-side overlap detection based on n-gram or token-level matching, memorization or extraction attacks that test whether benchmark content can be reproduced, perplexity-based methods that treat unusually low likelihood as evidence of prior exposure, and membership inference attacks (MIA) such as LiRA that estimate whether a sample was included in training \cite{brown2020gpt3,openai2023gpt4,carlini2021extracting,carlini2023quantifying,dekoninck2024constat,shi2024detecting,oren2024proving,carlini2022lira}. 

Despite the rich literature on contamination auditing in LLMs, these methods do not transfer directly to TSFMs. Overlap-based and extraction-based approaches rely on discrete tokens or verbatim matching, whereas time series are continuous values and the same underlying signal may reappear after rescaling, resampling, rewindowing, or renaming, making exact matching both brittle and incomplete \cite{yang2023rephrased,jiang2024investigating,meyer2025rethinking}. Perplexity- or loss-based criteria are also especially unreliable in time series, because many datasets are intrinsically easy to predict: smooth trends or strong seasonality can produce low error even without prior exposure, so static low loss alone is not a valid proxy for contamination \cite{dekoninck2024constat,jiang2024investigating,sainz2023nlp,dong2024generalization}. Membership inference methods face a different mismatch: they are typically designed for sample-level membership and often require shadow model calibration, while TSFM contamination is more naturally a dataset-level phenomenon and full TSFM pretraining is prohibitively expensive to reproduce \cite{shokri2017membership,carlini2022lira}. 

\paragraph{Our Positioning.}
Our work addresses this gap by formulating contamination auditing for TSFMs in a realistic setting where the auditor can probe model behavior and run limited fine tuning, but cannot inspect the full pretraining corpus or reproduce pretraining at scale. Our method \textsc{TSFMAudit} uses adaptation dynamics, rather than static loss alone, as the primary signal, following the broader observation that training dynamics can reveal dataset and model behavior \cite{swayamdipta2020dataset,frankle2020early,zhang2017understanding}. To the best of our knowledge, the closest related work is \citet{dekoninck2024constat}, which studies performance-based contamination detection for discrete language benchmarks, whereas \textsc{TSFMAudit} focuses on continuous time series and does not require shadow model calibration.

\section{Problem Formulation}
\label{sec:define}
In this section, we formalize the contamination auditing problem for multivariate TSFMs. We introduce the forecasting setup, a notion of pretraining contamination, and the auditing objective.

\paragraph{Setup}
We consider a time series forecasting setting with lookback length $L$, forecast horizon $H$, and number of variables (channels) $P$.
Let
\(
f_{\bm{\theta}} : \mathbb{R}^{L \times P} \to \mathbb{R}^{H \times P}
\)
denote a candidate TSFM with parameter
\(
\bm{\theta}.
\)
For sample $i$, let
\(
\mathbf{H}_i \in \mathbb{R}^{L \times P}
\)
be the historical input and
\(
\mathbf{Y}_i \in \mathbb{R}^{H \times P}
\)
be the future target. 
Following the supervised forecasting formulation, we define the $i$-th sample and an audited dataset with $n$ samples by
\[
\mathbf{Z}_i := (\mathbf{H}_i;\mathbf{Y}_i) \in \mathbb{R}^{(L+H)\times P}, \qquad \mathcal{D} := \{\mathbf{Z}_i\}_{i=1}^{n}.
\]

\paragraph{Contamination}

Let
\(
\mathcal{D}^{\mathrm{pt}}
=
\bigcup_{j=1}^{J}\mathcal{D}^{\mathrm{pt}}_j,
\)
denote the pretraining corpus associated with the candidate model
\(f_{\bm{\theta}}\), where \(J\) is the number of pretraining sources.
Here, \(\mathcal{D}^{\mathrm{pt}}\) is understood as the collection of
pretraining source time series.
For each audited sample
\(
\mathbf{Z}_i,
\)
let
\(
\mathbf{X}_i
\)
denote the underlying source time series from which
\(
\mathbf{Z}_i
\)
is extracted, for example through windowing or preprocessing.
Accordingly, we define the sample-level contamination indicator for sample
\(\mathbf{Z}_i\) as
\[
c_i=c\!\left(\mathbf{Z}_i;\mathcal{D}^{\mathrm{pt}}\right)
:=
\begin{cases}
1, & \mathbf{X}_i \in \mathcal{D}^{\mathrm{pt}}, \\[4pt]
0, & \mathbf{X}_i \notin \mathcal{D}^{\mathrm{pt}}.
\end{cases}
\]

The corresponding dataset-level contamination ratio is defined as
\[
C\!\left(\mathcal{D},\mathcal{D}^{\mathrm{pt}}\right)
:=
\frac{1}{n}\sum_{i=1}^{n} c_i
\in [0,1].
\]

Under this definition,
\(
C\!\left(\mathcal{D},\mathcal{D}^{\mathrm{pt}}\right)=0
\)
means that no audited sample is derived from a source time series that
appears in the pretraining corpus, while
\(
C\!\left(\mathcal{D},\mathcal{D}^{\mathrm{pt}}\right)=1
\)
means that all audited samples are derived from source time series
contained in the pretraining corpus.

\paragraph{Objective}

Given a candidate model $f_{\bm{\theta}}$ and an audited dataset $\mathcal{D}$, the goal of contamination auditing is to estimate whether $\mathcal{D}$ has been exposed during pretraining. More precisely, we seek a scoring function
\[
S\!\left(f_{\bm{\theta}},\mathcal{D}\right) \in [0,1]
\]
that quantifies the contamination risk of the pair $\left(f_{\bm{\theta}}, \mathcal{D}\right)$, where larger values indicate stronger evidence of pretraining contamination.
A binary decision can then be obtained by thresholding the score:
\[
\widehat{S}\!\left(f_{\bm{\theta}},\mathcal{D}\right)
:=
\begin{cases}
1, & S\!\left(f_{\bm{\theta}},\mathcal{D}\right)> \tau, \\[4pt]
0, & S\!\left(f_{\bm{\theta}},\mathcal{D}\right)\le \tau,
\end{cases}
\]
where $\tau \in [0,1]$ is a decision threshold, and $\widehat{S}=1$ denotes ``contaminated''.
Under the idealized formulation, one may directly take
\(
S\!\left(f_{\bm{\theta}},\mathcal{D}\right)
=
C\!\left(\mathcal{D},\mathcal{D}^{\mathrm{pt}}\right).
\)In practice, $C$ is generally unobservable because the pretraining corpus is only partially documented. We therefore work with an observable proxy $\widetilde{C}\!\left(f_{\bm{\theta}},\mathcal{D}\right)\in\{0,1\}$ derived from official training-source documentation; see Appendix~\ref{sec:appendix_labels} for the full definition and Appendix~\ref{sec:appendix_semantics} for transformed duplicate semantics.


\section{\textsc{TSFMAudit}: Debiased Adaptation Efficiency}
\label{sec:method}

\begin{figure}[t]
    \centering
    \includegraphics[width=\linewidth]{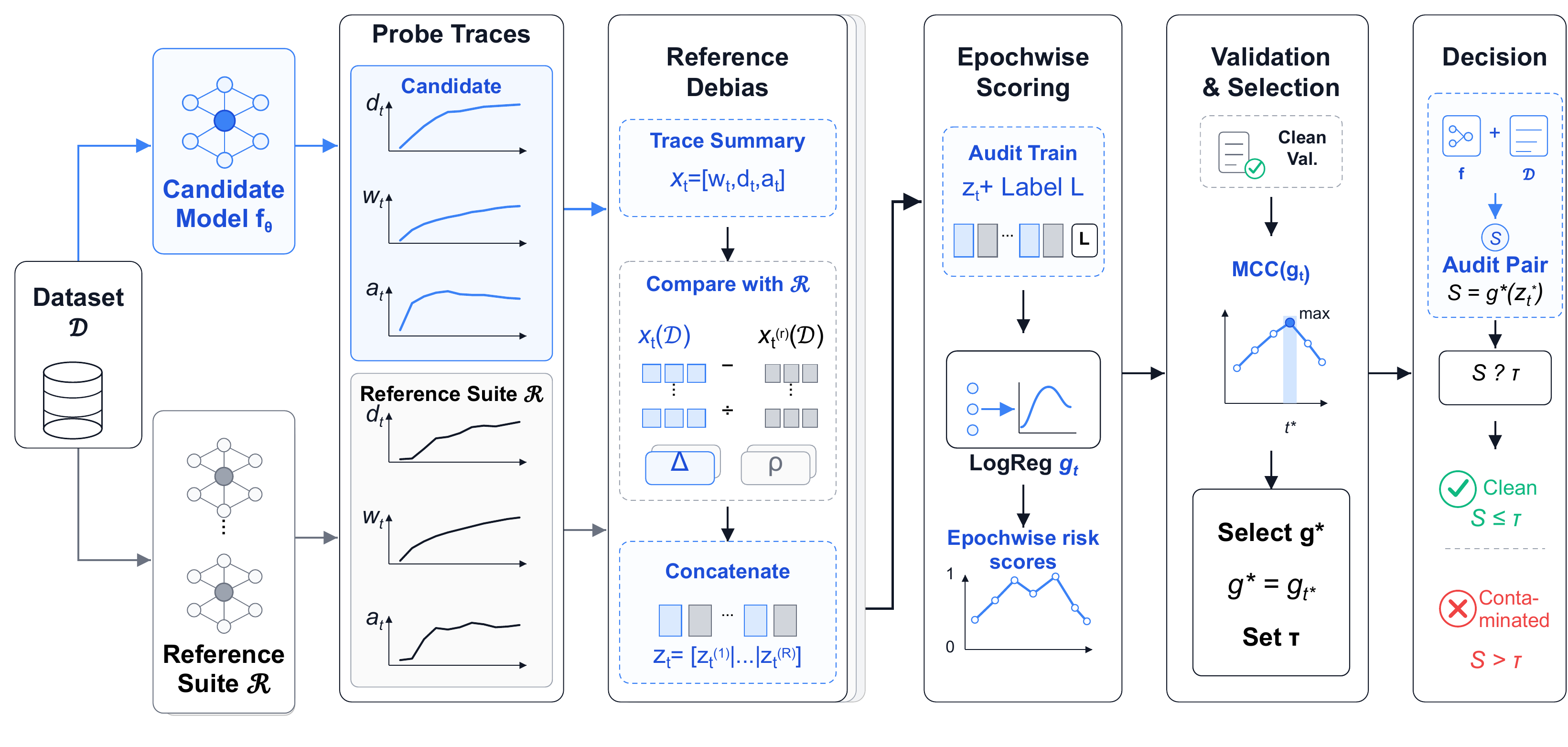}
    \caption{
    Overview of \textsc{TSFMAudit}.
    A candidate model and a reference suite are probed on \(\mathcal{D}\) for
    \(T_{\mathrm{probe}}\) gradient epochs under the same fine tuning protocol.
    The resulting traces are summarized into adaptation features, transformed into
    reference-based debiased features, and mapped to a scalar contamination-risk
    score \(S(f_{\bm{\theta}},\mathcal{D})\).
    }
    \label{fig:pipeline}
\end{figure}

We now describe \textsc{TSFMAudit}, which constructs the score
$S\left(f_{\bm{\theta}},\mathcal{D}\right)$ by evaluating how efficiently a candidate model adapts to
$\mathcal{D}$ during fine tuning.
Figure~\ref{fig:pipeline} provides an overview of the full pipeline.
The design of \textsc{TSFMAudit} is guided by two core observations.

\begin{wrapfigure}[12]{r}{7.5cm}
\vspace{-0.5cm}
    \centering
    \includegraphics[width=\linewidth]{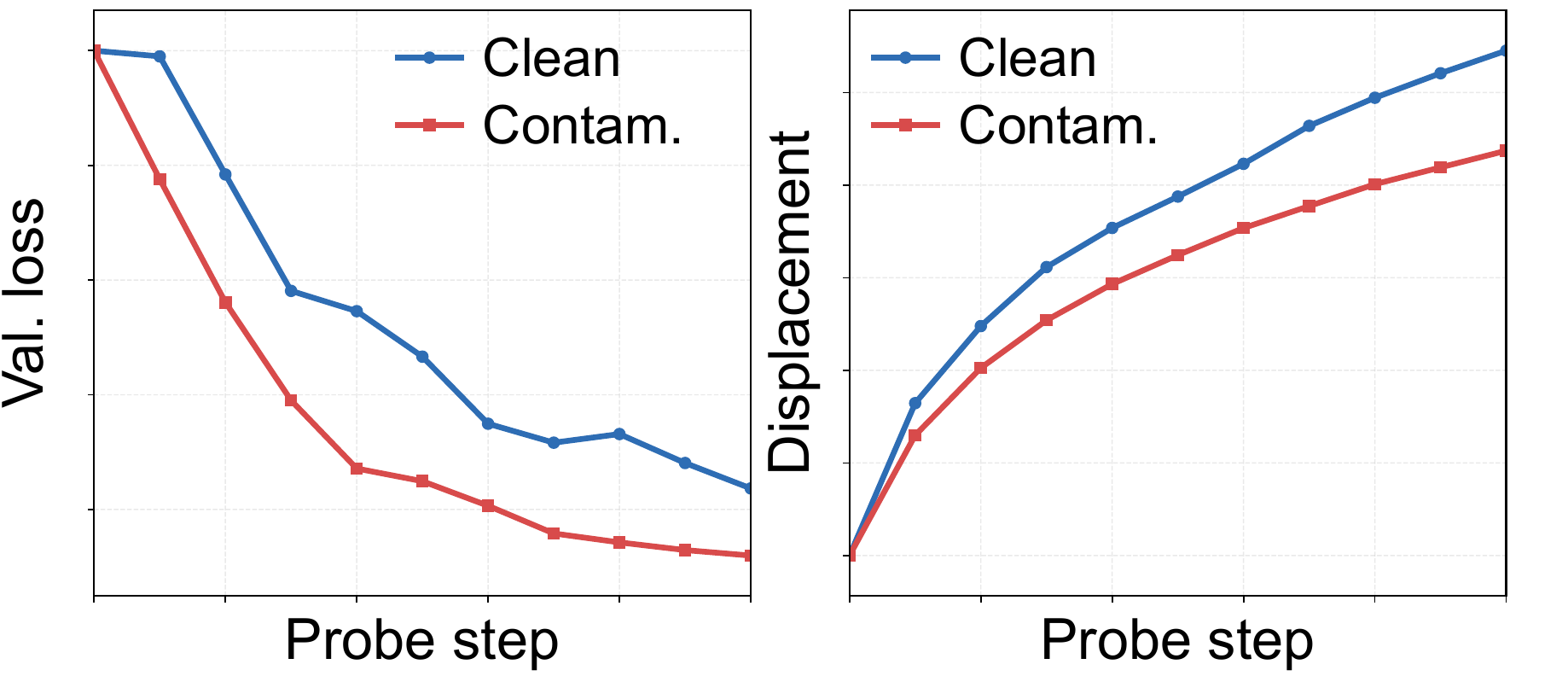}
    \caption{
    Schematic probe-time adaptation behavior.
    }
    \label{fig:trace}
\end{wrapfigure}
\paragraph{Observation 1: Contamination can appear as unusually efficient adaptation.}
If a model has already encoded predictive structure specific to
$\mathcal{D}$ during pretraining, it may require less adaptation when
subsequently fine tuning on $\mathcal{D}$.
In such cases, the model can exhibit faster loss reduction with relatively little backbone movement (as shown in Figure~\ref{fig:trace}).
We treat this contrast as behavioral evidence of possible contamination.
This interpretation is motivated by prior work on neural network memorization
\cite{zhang2017understanding,feldman2020neural} and recent detection (audit) methods
that use fine tuning induced change as evidence of leakage \cite{kds2025}.


\paragraph{Observation 2: Dataset difficulty can bias adaptation behavior.}
Fast adaptation is not unique to contaminated models: a clean model may adapt just as quickly if $\mathcal{D}$ is easy.
To reduce this difficulty-induced bias, we compare the candidate model against a reference suite $\mathcal{R}$ consisting of models with comparable
architecture and scale but no exposure to $\mathcal{D}$.
If contamination is substantial, the candidate may exhibit consistently higher adaptation efficiency than these references across varying difficulty levels.
This strategy follows the high-level motivation of reference-based tests such
as ConStat \cite{dekoninck2024constat}.

\subsection{Adaptation Traces for Auditing}

We construct the auditing signal by probing the candidate model
$f_{\bm{\theta}}$ and a reference suite
\(
\mathcal{R}=\left\{f_{\bm{\theta}^{(r)}}\right\}_{r=1}^{|\mathcal{R}|}
\)
under a shared fine tuning protocol on the audited dataset
$\mathcal{D}$.

The probing phase runs for $T_{\mathrm{probe}}$ gradient epochs fine-tuning
$f_{\bm{\theta}}$ on $\mathcal{D}$.
The resulting optimization trace is
\[
\mathcal{T}\!\left(f_{\bm{\theta}},\mathcal{D}\right)
=
\left\{\ell_t,\; w_t\right\}_{t=0}^{T_{\mathrm{probe}}},
\qquad
\ell_t
:=
\frac{1}{nHP}
\sum_{i=1}^{n}
\left\|
f_{\bm{\theta},t}(\mathbf{H}_i)-\mathbf{Y}_i
\right\|_F^2,
\qquad
w_t
:=
\left\|
\bm{\theta}^{\mathrm{bb}}_t-\bm{\theta}^{\mathrm{bb}}_0
\right\|_2,
\]
where $\ell_t$ is the loss at epoch $t$, and $w_t$ is the cumulative $\ell_2$ displacement of the backbone parameters from their initial values.
For each reference model $f_{\bm{\theta}^{(r)}} \in \mathcal{R}$, we define the corresponding probe trace $\mathcal{T}^{(r)}=\{\ell_t^{(r)},w_t^{(r)}\}_{t=0}^{T_{\mathrm{probe}}}$ analogously, with
\[
\ell_t^{(r)}
:=
\frac{1}{nHP}
\sum_{i=1}^{n}
\left\|
f_{\bm{\theta}_t^{(r)}}(\mathbf{H}_i)-\mathbf{Y}_i
\right\|_F^2,
\qquad
w_t^{(r)}
:=
\left\|
\bm{\theta}^{(r)}_t-\bm{\theta}^{(r)}_0
\right\|_2.
\]

From the candidate trace, we compute at each epoch $t$ the relative loss drop and the adaptation efficiency for candidate and reference models
\[
d_t
:=
\frac{\ell_0-\ell_t}{\ell_0+\varepsilon_{\ell}},
\qquad
a_t
:=
\frac{d_t}{w_t+\varepsilon_{\mathrm{ae}}},
\qquad
d_t^{(r)}
:=
\frac{\ell_0^{(r)}-\ell_t^{(r)}}{\ell_0^{(r)}+\varepsilon_{\ell}},
\qquad
a_t^{(r)}
:=
\frac{d_t^{(r)}}{w_t^{(r)}+\varepsilon_{\mathrm{ae}}},
\]
where $\varepsilon_{\ell}>0$ and $\varepsilon_{\mathrm{ae}}>0$ are
numerical stabilizers. A large loss drop $d_t$ together with a small backbone displacement $w_t$
yields a high $a_t$, which we interpret as behavioral evidence of possible
contamination.

At a fixed set of checkpoints,
we summarize the candidate and reference traces by
\[
\mathbf{x}(\mathcal{D})
=
\left[w_t,\; d_t,\; a_t\right]_{ 1\leq t\leq T_{\mathrm{probe}}},
\qquad
\mathbf{x}^{(r)}(\mathcal{D})
=
\left[w_t^{(r)},\; d_t^{(r)},\; a_t^{(r)}\right]_{1\leq t\leq T_{\mathrm{probe}}}.
\]


\subsection{Debiasing with Reference Models}
\label{sec:relative_debiasing}

The trace summaries introduced above are affected by dataset-specific factors that are shared across models, most notably the intrinsic difficulty of $\mathcal{D}$.
To reduce this source of bias, \textsc{TSFMAudit} compares the candidate model against each reference model in $\mathcal{R}$ under the same probing protocol.
Accordingly, for each epoch $t$ and each reference $r$, we define the elementwise difference and ratio
\[
\bm{\Delta}_t^{(r)}(\mathcal{D}) = 
\mathbf{x}_t(\mathcal{D})-\mathbf{x}_t^{(r)}(\mathcal{D}) \in \mathbb{R}^{3},
\qquad
\bm{\rho}_t^{(r)}(\mathcal{D}) = 
\mathbf{x}_t(\mathcal{D})
\oslash
\left(
\mathbf{x}_t^{(r)}(\mathcal{D})+\varepsilon_{\mathrm{ref}}\mathbf{1}
\right) \in \mathbb{R}^{3},
\]
where $\oslash$ denotes elementwise division and $\varepsilon_{\mathrm{ref}}>0$ is a numerical stabilizer.

For each probe epoch $t$, let
\(
g_t:\mathbb{R}^{6|\mathcal{R}|}\rightarrow[0,1]
\label{eq:epochwise_scorer}
\)
denote a lightweight scorer (logistic regression more precisely) applied to the reference-debiased representation obtained by concatenating
$\{\bm{\Delta}_t^{(r)}(\mathcal{D}), \bm{\rho}_t^{(r)}(\mathcal{D})\}_{r=1}^{|\mathcal{R}|}$.
The scorer $g_t$ is fitted on an audit training set whose labels are the observable proxy indicators
\(
\widetilde{C}\!\left(f_{\bm{\theta}},\mathcal{D}\right)\in\{0,1\}.
\)

Therefore, $g_t$ outputs a contamination probability conditioned on
the relative adaptation behavior at probe epoch $t$.

\subsection{Calibration and Decision}
\label{sec:calibration_decision}
Among the epochwise scorers $\{g_t\}_{t=1}^{T_{\mathrm{probe}}}$, we select the
final scorer $g^\ast$ using a validation set with known contamination labels.
Specifically, we evaluate each $g_t$ on the validation set and choose the probe epoch with the highest Matthews correlation coefficient (MCC)~\cite{Matthews1975T4Lysozyme}.

To convert this score into a binary decision, we adopt a conservative
``do-not-accuse-the-innocent'' principle.
That is, when setting the decision threshold, we prioritize avoiding false accusations on clean datasets.
Let $\mathcal{V}_{\mathrm{clean}}$ denote the collection of validation datasets known to be clean, i.e., satisfying
$C\!\left(\mathcal{D},\mathcal{D}^{\mathrm{pt}}\right)=0$.
We then choose the threshold $\tau$ as the largest value such that no clean validation dataset is classified as contaminated:
\[
\tau
=
\max_{\mathcal{D}\in\mathcal{V}_{\mathrm{clean}}}
g^\ast\!\{\bm{\Delta}_t^{(r)}(\mathcal{D}), \bm{\rho}_t^{(r)}(\mathcal{D})\}_{r=1}^{|\mathcal{R}|}.
\]
The final binary audit decision is
\[
\widehat{S}\!\left(f_{\bm{\theta}},\mathcal{D}\right)
=
\mathbb{I}\!\left[
S\!\left(f_{\bm{\theta}},\mathcal{D}\right) > \tau
\right]
=
\mathbb{I}\!\left[
g^\ast\!\{\bm{\Delta}_t^{(r)}(\mathcal{D}), \bm{\rho}_t^{(r)}(\mathcal{D})\}_{r=1}^{|\mathcal{R}|} > \tau
\right].
\]
This conservative calibration may sacrifice some sensitivity, but it enforces a strict no-false-positive criterion on the clean validation set. Under this rule no clean validation pair is flagged as contaminated, which instantiates the FP-$0$ calibration protocol of Section~\ref{sec:appendix_protocol}.

\section{Experiments}
\label{sec:experiments}

Our experiments are designed to answer four questions:
(Q1) Whether \textsc{TSFMAudit} can effectively identify contamination, and how it performs relative to competitive baselines.
(Q2) Whether reference models provide a debiasing effect in contamination detection.
(Q3) Whether each proposed feature makes a meaningful contribution to the final performance.
(Q4) Whether \textsc{TSFMAudit} remains effective in practical deployment on a fresh external benchmark.

\subsection{Experimental Setup}
\label{sec:setup}

\begin{wraptable}[11]{r}{8.3cm}
\vspace{-0.5cm}
    \centering
    \caption{Class distribution per candidate TSFM. 
    }
    \label{tab:class_balance}
    \begin{tabular}{lrrrl}
        \toprule
        \textbf{Candidate} & \textbf{Clean} & \textbf{Leaked} & \textbf{Total} & \textbf{Prevalence} \\
        \midrule
        Chronos & 118 & 19 & 137 & 13.9\% \\
        TiRex & 118 & 19 & 137 & 13.9\% \\
        TimesFM2.0 & 125 & 12 & 137 & 8.8\% \\
        Kairos & 20 & 117 & 137 & 85.4\% \\
        Moirai2 & 20 & 117 & 137 & 85.4\% \\
        Moirai1 & 5 & 132 & 137 & 96.4\% \\
        \bottomrule
    \end{tabular}
\end{wraptable}

\paragraph{Candidates and datasets.}
We audit six TSFMs: Chronos~\cite{ansari2024chronos}, TiRex~\cite{auer2025tirex}, Moirai1/2~\cite{woo2024moirai}, TimesFM2.0~\cite{das2024timesfm}, and Kairos~\cite{feng2025kairos}.
The dataset pool combines benchmark datasets from GIFT-Eval and the GIFT-Eval Pretrain corpus~\cite{aksu2024gift}, covering both evaluation-side and pretraining-side catalogs used by recent TSFMs. We retain datasets with $\geq$30 series, each $\geq$152 samples, and subsample to $\leq$50 series per dataset; this yields 137 audited datasets.
Details are in Appendix~\ref{sec:appendix_datasets}.

Ground truth contamination labels are unavailable.
We construct proxy labels by comparing each TSFM's documented training sources against benchmark metadata.
Table~\ref{tab:class_balance} summarizes class balance.

\paragraph{Baselines.}
We compare \textsc{TSFMAudit} to static-loss and MIA-inspired baselines, all evaluated under the same FP-$k$ protocol:
\begin{itemize}[leftmargin=1.5em, itemsep=1pt, topsep=2pt]
    \item \textbf{Raw Loss}: Candidate pre-probe loss $\ell_0$.
    
    \item \textbf{Loss Drop Rate}: Candidate relative loss reduction after probing,
    $
    d_{T_{\mathrm{probe}}}
    =
    \frac{\ell_0-\ell_{T_{\mathrm{probe}}}}{\ell_0+\varepsilon_\ell}.
    $
    
    \item \textbf{LiRA Ratio}: LiRA-inspired~\cite{carlini2022lira} zero-probe loss ratio
    $
    \ell_0 / \ell_0^{(r)},
    $
    evaluated separately for each reference model $r$.
    
    \item \textbf{TS MinK FFT}: Min-K\%~\cite{shi2024detecting} style frequency-domain score $S_{\mathrm{FFT}}(f_{\bm{\theta}}, \mathcal{D})$ computed on forecast residuals across all six candidate TSFMs; see Appendix~\ref{sec:appendix_fft_baseline}.
    
    \item \textbf{Dyn no Ref}: Candidate-only dynamics baseline using the probe summary without reference-based debiasing.
\end{itemize}

\paragraph{Reference models.}
To reduce bias induced by intrinsic dataset difficulty, we instantiate the reference suite
$
\mathcal{R}=\{f_{\bm{\theta}^{(r)}}\}_{r=1}^{|\mathcal{R}|}
$
with four models: ScratchCNN, ScratchTransformer, Stat, and VisionTS. None of them has exposure to time-series pretraining, so they provide reference baselines for dataset difficulty and adaptation speed under downstream fine tuning.
ScratchCNN and ScratchTransformer are lightweight forecasters initialized randomly and trained from scratch, with convolutional and self-attention inductive biases, respectively.
Stat aggregates predictions from methods such as ETS~\cite{Hyndman2021FPP3} and serves as a non-parametric difficulty anchor.
VisionTS~\citep{chen2025visionts} is a TSFM whose representations are derived from a masked autoencoder trained on natural images, providing a pretrained reference with no time-series pretraining exposure.

\paragraph{All-reference configurations.}
For compactness, we use two roman-text labels for the multi-reference variants.
All Ref denotes the configuration that concatenates the difference and ratio features
$\{\bm{\Delta}_t^{(r)}(\mathcal{D}),\bm{\rho}_t^{(r)}(\mathcal{D})\}_{r=1}^{|\mathcal{R}|}$
over all reference models and fits the logistic-regression scorer $g_t$.
All Ref+Det uses the same all-reference representation and additionally appends the direct candidate dynamics (Det), namely the candidate trace summary
$\mathbf{x}(\mathcal{D})=[w_t,d_t,a_t]_{1\leq t\leq T_{\mathrm{probe}}}$.
We use the labels All Ref and All Ref+Det consistently in the text and tables.

\paragraph{Evaluation metrics.}
We report Matthews Correlation Coefficient (MCC)~\cite{Matthews1975T4Lysozyme}, Macro-F1, AUROC, and balanced accuracy 
(Bal.Acc) to capture both ranking quality and thresholded decision performance under class 
imbalance. Denoting true/false positives and negatives as TP, TN, FP, FN:
\[
\mathrm{MCC} = \frac{\mathrm{TP}\cdot\mathrm{TN}-\mathrm{FP}\cdot\mathrm{FN}}
{\sqrt{(\mathrm{TP}+\mathrm{FP})(\mathrm{TP}+\mathrm{FN})(\mathrm{TN}+\mathrm{FP})
(\mathrm{TN}+\mathrm{FN})}}, 
\mathrm{Bal.Acc} = \tfrac{1}{2}\!\left(
\tfrac{\mathrm{TP}}{\mathrm{TP}+\mathrm{FN}}+\tfrac{\mathrm{TN}}{\mathrm{TN}+\mathrm{FP}}
\right).
\]
Macro-F1 is the unweighted mean of per-class F1 scores; AUROC is the area under the receiver 
operating characteristic curve, which plots the true positive rate against the false positive 
rate over all decision thresholds. Higher is better for all metrics.

\subsection{Main Results: Can \textsc{TSFMAudit} Identify Contamination?}
\label{sec:main_results}

We begin by answering Q1, Table~\ref{tab:main_results} reports the main results.
\begin{table*}[t]
\centering
\scriptsize
\setlength{\tabcolsep}{3.5pt}
\begin{tabular*}{\textwidth}{@{\extracolsep{\fill}}llccc@{}}
\toprule
\textbf{Family} & \textbf{Method} & \textbf{MCC} $\uparrow$ & \textbf{Macro-F1} $\uparrow$ & \textbf{Bal.Acc} $\uparrow$ \\
\midrule
\multirow{2}{*}{\textbf{Static baselines}}
& Raw Loss ($\ell_0$) & 0.044 $\pm$ 0.123 & 0.342 $\pm$ 0.255 & 0.528 $\pm$ 0.054 \\
& Loss Drop Rate ($d_{T_{\mathrm{probe}}}$) & 0.055 $\pm$ 0.070 & 0.346 $\pm$ 0.217 & 0.522 $\pm$ 0.061 \\
\midrule
\multirow{4}{*}{\textbf{LiRA-inspired}}
& LiRA Ratio (ScratchCNN) & $-0.005 \pm 0.039$ & 0.332 $\pm$ 0.191 & 0.517 $\pm$ 0.045 \\
& LiRA Ratio (ScratchTransformer) & $-0.031 \pm 0.024$ & 0.299 $\pm$ 0.215 & 0.487 $\pm$ 0.010 \\
& LiRA Ratio (Stat) & $-0.001 \pm 0.049$ & 0.292 $\pm$ 0.242 & 0.492 $\pm$ 0.021 \\
& LiRA Ratio (VisionTS) & 0.012 $\pm$ 0.030 & 0.334 $\pm$ 0.184 & 0.501 $\pm$ 0.037 \\
\midrule
\multirow{3}{*}{\textbf{Frequency baseline}}
& TS MinK FFT ($K=10\%$) & $0.005 \pm 0.116$ & $0.324 \pm 0.247$ & 0.490 $\pm$ 0.052 \\
& TS MinK FFT ($K=20\%$) & $0.004 \pm 0.117$ & $0.323 \pm 0.249$ & 0.494 $\pm$ 0.044 \\
& TS MinK FFT ($K=30\%$) & $0.015 \pm 0.113$ & $0.339 \pm 0.244$ & 0.512 $\pm$ 0.030 \\
\midrule
\textbf{Dynamics baseline}
& Candidate-only dynamics (No Ref) & 0.032 $\pm$ 0.061 & 0.411 $\pm$ 0.113 & 0.539 $\pm$ 0.056 \\
\midrule
\multirow{6}{*}{\textbf{\textsc{TSFMAudit} (ours)}}
& + ScratchCNN & 0.087 $\pm$ 0.125 & 0.453 $\pm$ 0.140 & 0.560 $\pm$ 0.078 \\
& + ScratchTransformer & 0.045 $\pm$ 0.046 & 0.420 $\pm$ 0.132 & 0.529 $\pm$ 0.020 \\
& + Stat & 0.046 $\pm$ 0.073 & 0.420 $\pm$ 0.128 & 0.531 $\pm$ 0.049 \\
& + VisionTS & 0.056 $\pm$ 0.070 & 0.458 $\pm$ 0.086 & 0.548 $\pm$ 0.054 \\
& All Ref+Det & 0.102 $\pm$ 0.116 & 0.510 $\pm$ 0.078 & 0.589 $\pm$ 0.085 \\
& All Ref & \textbf{0.125 $\pm$ 0.090} & \textbf{0.521 $\pm$ 0.077} & \textbf{0.603 $\pm$ 0.063} \\
\bottomrule
\end{tabular*}
\caption{Main effectiveness comparison, higher is better.}
\label{tab:main_results}
\end{table*}
The overall pattern is clear: static-loss baselines and LiRA-inspired ratios are weak detectors.
Raw loss and loss-drop rate yield only small positive MCC values, while all LiRA variants remain near zero or even negative in MCC.
This suggests that contamination in TSFMs cannot be reliably identified from pre-probe loss alone, nor from a simple reference loss ratio.
The frequency baseline also fails to provide a stable signal: although it attains moderate Macro-F1 for some $K$, its MCC is close to zero ($\le 0.015$) with high variance, indicating poor agreement with the contamination labels under thresholded decision making.
Probe dynamics provide a substantially more useful signal. Even without any reference model, the dynamics baseline already improves Macro-F1.
This result supports our central intuition that contamination is reflected not only in how well a model predicts before probing, but also in how efficiently it adapts during the fine tuning phase.
The full \textsc{TSFMAudit} method improves further once reference-based debiasing is introduced.
Across the single-reference choices and the two all-reference combinations, all reference-based variants outperform the candidate-only dynamics baseline in Macro-F1 and MCC.
The strongest setting is All Ref, which uses all reference models through the debiased difference and ratio features.
\begin{figure}[tb]
\centering
\includegraphics[width=\linewidth]{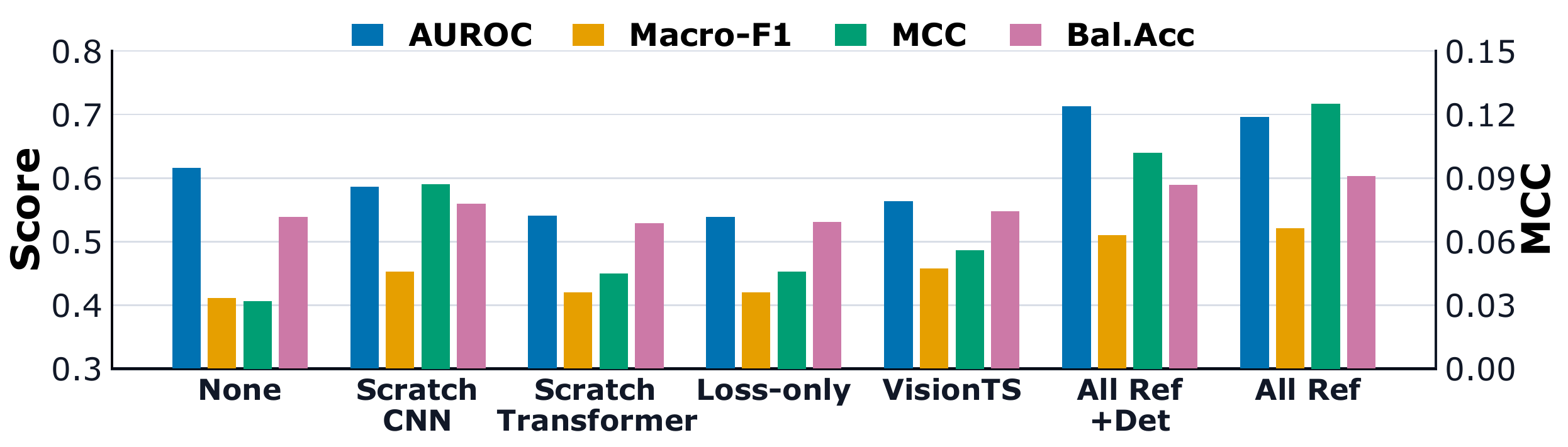}
\caption{Reference model ablation. Higher is better.}
\label{fig:ref_ablation}
\end{figure}

Taken together, these results provide a positive answer to Q1. \textsc{TSFMAudit} is able to distinguish contaminated datasets from clean ones more effectively than static-loss, LiRA-inspired, and frequency domain baselines.
At the same time, the absolute MCC values remain modest, which highlights that this is a challenging detection problem under conservative FP-$k$ calibration and noisy proxy labels.

\subsection{Reference-Based Debiasing}
\label{sec:ref_debiasing}

We next answer Q2, Figure~\ref{fig:ref_ablation} compares different single-reference choices and all-reference combinations.
Table~\ref{tab:debias_ablation} then ablates how the candidate and reference traces are combined.

The first conclusion is that reference comparison provides a useful debiasing effect.
Compared with the no-reference setting, all reference-based variants improve MCC and Macro-F1, although the gains vary by reference construction.
The all-reference combinations provide the strongest aggregate results: All Ref achieves the best Macro-F1, MCC, and balanced accuracy, while All Ref+Det gives the highest AUROC.
The second conclusion is that reference-based debiasing improves thresholded decision quality, and the all-reference combinations further improve ranking quality.
Under our FP-$k$ protocol, this is the more relevant behavior: the goal is not only to rank datasets, but to make accurate audit decisions after conservative threshold calibration. From this perspective, reference models are beneficial overall.

\begin{wraptable}[9]{r}{8.3cm}
\vspace{-0.5cm}
\centering
\small
\setlength{\tabcolsep}{6pt}
\begin{tabular}{lcc}
\toprule
\textbf{Debiasing mode} & \textbf{MCC} $\uparrow$ & \textbf{Macro-F1} $\uparrow$ \\
\midrule
$X_c$ only (no debiasing) & 0.032 $\pm$ 0.061 & 0.411 $\pm$ 0.113 \\
$X_{\mathrm{diff}}$ ($X_c - X_r$) & 0.038 $\pm$ 0.058 & 0.425 $\pm$ 0.095 \\
$X_{\mathrm{ratio}}$ ($X_c / X_r$) & 0.048 $\pm$ 0.065 & 0.442 $\pm$ 0.088 \\
$X_{\mathrm{inter}}$ ($X_c \odot X_r$) & 0.025 $\pm$ 0.048 & 0.398 $\pm$ 0.102 \\
Full concat (all) & \textbf{0.056 $\pm$ 0.070} & \textbf{0.458 $\pm$ 0.086} \\
\bottomrule
\end{tabular}
\caption{Ablation of debiasing representations.}
\label{tab:debias_ablation}
\end{wraptable}
Table~\ref{tab:debias_ablation} shows how the debiasing effect arises from relative features.
Using only candidate features $X_c$ is the weakest setting among the competitive variants.
Once we introduce relative candidate-reference features, performance improves.
Among single debiasing modes, ratio features $X_{\mathrm{ratio}}$ perform best, outperforming both difference features $X_{\mathrm{diff}}$ and interaction features $X_{\mathrm{inter}}$.
This indicates that scale-normalized comparisons between candidate and reference behavior are more informative than raw additive or multiplicative relations alone.
The best overall performance is obtained by concatenating all debiased views.
This result suggests that the different transformations capture complementary aspects of the candidate-reference relationship.

Taken together, these results give a clear answer to Q2. Yes, reference models provide a debiasing effect. They help by converting probe dynamics into relative candidate-reference signals, which reduce the influence of dataset easiness on the audit score.

\subsection{Feature Ablation: Are the Proposed Features Useful?}
\label{sec:feature_ablation}

We next answer Q3 by ablating the probe-time features used in \textsc{TSFMAudit}.
Table~\ref{tab:feature_ablation} compares auditor built from a single feature family against the full dynamics representation.
The main result is that no single feature is sufficient on its own. All individual features yield weaker performance than the full dynamics representation.
This shows that the contamination signal is distributed across multiple aspects of probe time behavior.

\begin{wraptable}[9]{r}{8.3cm}
\vspace{-0.5cm}
\centering
\setlength{\tabcolsep}{3.5pt}
\begin{tabular}{lcc}
\toprule
\textbf{Feature} & \textbf{MCC} $\uparrow$ & \textbf{Macro-F1} $\uparrow$ \\
\midrule
Loss & 0.021 $\pm$ 0.045 & 0.385 $\pm$ 0.098 \\
Loss drop rate & 0.028 $\pm$ 0.052 & 0.392 $\pm$ 0.105 \\
Adaptation efficiency & \textbf{0.045 $\pm$ 0.068} & \textbf{0.418 $\pm$ 0.112} \\
Parameter displacement & 0.018 $\pm$ 0.041 & 0.378 $\pm$ 0.095 \\
Full dynamics (all) & \textbf{0.056 $\pm$ 0.070} & \textbf{0.458 $\pm$ 0.086} \\
\bottomrule
\end{tabular}
\caption{Feature ablation of \textsc{TSFMAudit}. Higher is better.}
\label{tab:feature_ablation}
\end{wraptable}

Among the individual features, adaptation efficiency is the strongest. It achieves the best standalone MCC and Macro-F1, which supports our core hypothesis: contaminated datasets tend to exhibit unusually efficient adaptation, combining faster loss reduction with smaller parameter movement.
Loss and loss-drop rate are weaker when used alone.
This suggests that loss-based quantities capture only part of the relevant behavior.
In particular, they do not distinguish whether a fast improvement comes from genuine prior exposure or simply from an intrinsically easy dataset.
Parameter displacement is the weakest standalone feature. It appears to play a complementary role.

Overall, these results give a clear answer to Q3.
Yes, the proposed features are useful, but not equally so. Adaptation efficiency is the strongest individual feature, while loss and displacement provide complementary information.
The best performance is obtained by combining all dynamics features, which indicates that \textsc{TSFMAudit} benefits from modeling contamination as a multidimensional probe time phenomenon.

\subsection{Practical Deployment on the TIME Benchmark}
\label{sec:time_negative_control}

\begin{wraptable}[12]{r}{0.4\linewidth}
\vspace{-0.5cm}
\centering
\setlength{\tabcolsep}{2.5pt}
\begin{tabular*}{\linewidth}{@{\extracolsep{\fill}}lrr@{}}
\toprule
\textbf{Candidate} & \textbf{Non-leak rate} & \textbf{95\% LB} \\
\midrule
Chronos    & $96.0\%$  & $87.9\%$ \\
Kairos     & $100.0\%$ & $94.2\%$ \\
Moirai1    & $92.0\%$  & $82.6\%$ \\
Moirai2    & $100.0\%$ & $94.2\%$ \\
TiRex      & $96.0\%$  & $87.9\%$ \\
TimesFM2.0 & $100.0\%$ & $94.2\%$ \\
\midrule
Pooled     & $97.33\%$ & $95.24\%$ \\
\bottomrule
\end{tabular*}
\caption{TIME Benchmark deployment check.}
\label{tab:time_negative_control}
\end{wraptable}

We finally answer Q4 by testing whether the audit can be used in a realistic setting where the target benchmark has no contamination labels available for fitting or threshold selection.
We use the TIME Benchmark~\cite{qiao2026time}, a recent external benchmark whose datasets were collected and organized after the release of the audited TSFMs. We therefore use TIME as a ground-truth negative-control benchmark with $100\%$ non-leakage labels for these candidates.
The All Ref scorer and FP-$0$ threshold are inherited from the main experiment and applied directly to TIME without retraining or recalibration; further protocol details are in Appendix~\ref{sec:appendix_time}.

Table~\ref{tab:time_negative_control} shows consistently high non-leakage rates on TIME.
The pooled one-sided lower bound remains above $95\%$, and the candidate-level lower bounds remain high despite using only $50$ TIME datasets per model.
Thus, under the fixed auditor, TIME does not show evidence of leakage for these TSFMs.
This result supports TIME's claim as a fresh benchmark built from recent data, and gives a positive answer to Q4: even without labels for the target datasets, the calibrated \textsc{TSFMAudit} pipeline can be applied directly and still produce useful audit decisions.

\section{Conclusion}
In this paper, we formalize the problem of pretraining contamination auditing for TSFMs and introduce \textsc{TSFMAudit}, a method based on probe time adaptation dynamics. 
Experiments on 6 TSFMs and 187 datasets show that this dynamic view provides a stronger contamination signal than static-loss, LiRA-inspired, and frequency-based baselines, and that reference-based debiasing further improves audit decisions. To the best of our knowledge, this work is the first systematic study of contamination auditing for TSFMs and the first dedicated framework tailored to this setting. While the task remains challenging, our results establish probe time dynamics as a promising foundation for reliable contamination auditing in time series.

\bibliographystyle{plainnat}
\bibliography{references}

\newpage
\appendix

\section{Benchmark Scope and Contamination Labels}
\label{sec:appendix_benchmark}

This appendix supports the main experimental benchmark used throughout the paper, which consists of six candidate TSFMs and 137 audited datasets after the uniform loading and filtering procedure in Section~\ref{sec:appendix_datasets}. This 137 dataset collection is the evaluated benchmark used in the quantitative experiments. Unless stated otherwise, all performance tables, figures, and candidate--dataset labels in this appendix refer to this same six-candidate, 137-dataset benchmark.

Throughout, we follow the notation of the main paper: $f_{\bm{\theta}}$ denotes the candidate TSFM, $\mathcal{D}$ the audited dataset, $\mathcal{D}^{\mathrm{pt}}$ the candidate's pretraining corpus, and $\mathcal{R}$ the reference suite. The basic audit unit is the candidate dataset pair $(f_{\bm{\theta}}, \mathcal{D})$.

\subsection{Dataset Pool and Filtering}
\label{sec:appendix_datasets}

The evaluated dataset pool combines (i) benchmark datasets from GIFT-Eval and (ii) datasets from GIFT-Eval Pretrain. After the uniform filtering and loading protocol described below, the final evaluated benchmark contains 137 datasets. This is the same dataset collection used in the quantitative experiments of the main paper.

\paragraph{Filtering rules.}
A dataset is retained only if it satisfies all of the following conditions:
\begin{itemize}
    \item it contains at least 30 valid series;
    \item each retained series contains at least $L+H=128+24=152$ time steps;
    \item nearly constant series with standard deviation below $10^{-6}$ are removed;
    \item at most 50 series are used per dataset in order to bound probe cost.
\end{itemize}
Datasets are converted to Arrow format for unified loading. This conversion is only an implementation convenience and does not alter the underlying series values.

\subsection{From Idealized Contamination to Observable Supervision}
\label{sec:appendix_labels}

Section~\ref{sec:define} defines the idealized contamination ratio
\[
C\!\left(\mathcal{D},\mathcal{D}^{\mathrm{pt}}\right) \in [0,1],
\]
which depends on exact sample overlap between the audited dataset and the hidden pretraining corpus. In realistic TSFM settings, however, the pretraining corpus $\mathcal{D}^{\mathrm{pt}}$ is usually only partially documented. As a result, the ground truth quantity $C\!\left(\mathcal{D},\mathcal{D}^{\mathrm{pt}}\right)$ is generally unobservable. In addition, the available documentary evidence depends on the candidate model under consideration. The same audited dataset may therefore receive different observable labels for different candidates. We therefore define the observable proxy at the level of the audit pair $\left(f_{\bm{\theta}}, \mathcal{D}\right)$, rather than at the level of the dataset alone.

For evaluation, we use the binary proxy label
\[
\widetilde{C}\!\left(f_{\bm{\theta}},\mathcal{D}\right)\in\{0,1\},
\]
defined for each audit pair $\left(f_{\bm{\theta}},\mathcal{D}\right)$. We set
\[
\widetilde{C}\!\left(f_{\bm{\theta}},\mathcal{D}\right)=1
\]
only when official sources associated with the candidate, including the paper, model card, dataset card, or official repository documentation, explicitly state that pretraining used the audited dataset. Otherwise, we set
\[
\widetilde{C}\!\left(f_{\bm{\theta}},\mathcal{D}\right)=0.
\]
A zero label should be interpreted as no documented evidence in official sources, rather than as evidence that overlap is absent.

This proxy label records documented exposure to the audited corpus. It should not be interpreted as direct evidence of benchmark test leakage. In particular, official documentation may establish shared provenance across renamed, rescaled, or repackaged variants even when exact sample overlap cannot be established from public documentation alone. Accordingly, the proxy label is intentionally broader than exact sample overlap, which serves in the main paper only as an idealized conceptual starting point. Appendix~\ref{sec:appendix_semantics} returns to this issue through a representative normalized-duplicate analysis.

To make this candidate dependence concrete, we briefly summarize how official sources describe the training data of each audited candidate:
\begin{itemize}
    \item \textbf{Chronos.} Official sources describe training on large collections of public datasets together with synthetic data~\citep{ansari2024chronos}.

    \item \textbf{TiRex.} The official model card lists \texttt{autogluon/chronos\_datasets} and \texttt{Salesforce/GiftEvalPretrain} among the training sources~\citep{auer2025tirex}.

    \item \textbf{Moirai1.} Official sources state that Moirai~1 is pretrained on LOTSA~\citep{woo2024moirai}.

    \item \textbf{Moirai2.} The official model card states that Moirai~2 uses a subset of the GIFT-Eval Pretrain and Train datasets as non-leaking historical context, together with additional synthetic and internal operational data~\citep{liu2025moirai}.

    \item \textbf{TimesFM2.0.} Official documentation states that TimesFM~2.0 uses the TimesFM~1.0 pretraining set together with a listed subset of LOTSA datasets~\citep{das2024timesfm}.

    \item \textbf{Kairos.} The official project page, repository, and paper describe Kairos as pretrained on the Predictability-Stratified Time Series corpus, which is stated to contain real-world observations from Chronos and Moirai together with synthetic data~\citep{feng2025kairos}.
\end{itemize}

\subsection{Contamination Semantics Beyond Exact Overlap}
\label{sec:appendix_semantics}

Section~\ref{sec:define} introduces exact sample overlap as an idealized conceptual starting point for contamination. It also notes that, in time series, contamination can be more subtle than literal reuse of identical values: the same underlying signal may reappear after rescaling, renaming, reindexing, or related preprocessing changes. We therefore examine a stricter semantic question: should contamination-relevant repetition be limited to exact overlap, or should transformed variants of the same underlying signal also be treated as leakage-relevant?

To probe this boundary directly, we construct a transformed-duplicate setting in which repeated datasets are first normalized in value and then passed through the same audit pipeline. All other components of the protocol are unchanged, including the candidate models, the reference suite, the repeated split construction, and the FP-0 calibration rule. If contamination-relevant repetition required literal value-level identity, then this normalization step would be expected to substantially weaken the audit signal.

Table~\ref{tab:normalized_duplicate_semantics} shows that this does not happen. Under the candidate-only baseline, the normalized setting still yields Pair-MICRO MCC $0.274$ and Macro-F1 $0.473$. Reference-based variants remain clearly above this baseline. For example, ScratchCNN reaches Pair-MICRO MCC $0.389$ and Macro-F1 $0.628$, while All Ref reaches Pair-MICRO MCC $0.415$ and Macro-F1 $0.642$. The same pattern also appears at the Model-MACRO level: the no-reference baseline attains MCC $0.101$ and Macro-F1 $0.434$, whereas All Ref reaches MCC $0.128$ and Macro-F1 $0.508$.

These results indicate that contamination-relevant repetition in TSFM auditing is not limited to exact raw-value overlap. When normalization preserves the underlying signal structure and provenance, the audit signal often remains detectable. This does not imply invariance to arbitrary preprocessing, nor does it collapse all transformed variants into the same contamination event. Rather, it shows that exact overlap is too narrow as the sole operational notion of leakage in time series. This is consistent with the broader proxy semantics adopted in Section~\ref{sec:appendix_labels}, where documented shared provenance can remain contamination-relevant even when public documentation does not establish literal sample identity.

\begin{table}[t]
  \caption{Representative results in the normalized-duplicate setting under the same FP-0 protocol and family-aware splits. Useful audit signal remains after value normalization, which suggests that contamination-relevant repetition is not limited to exact raw-value overlap.}
  \label{tab:normalized_duplicate_semantics}
  \centering
  \small
  \begin{tabular}{lcccc}
    \toprule
    \textbf{Reference} & \textbf{Model-MACRO} & \textbf{Model-MACRO} & \textbf{Pair-MICRO} & \textbf{Pair-MICRO} \\
     & \textbf{Macro-F1} $\uparrow$ & \textbf{MCC} $\uparrow$ & \textbf{Macro-F1} $\uparrow$ & \textbf{MCC} $\uparrow$ \\
    \midrule
    None         & 0.434$\pm$0.175 & 0.101$\pm$0.124 & 0.473 & 0.274 \\
    ScratchCNN   & 0.492$\pm$0.095 & 0.101$\pm$0.108 & 0.628 & 0.389 \\
    Stat         & 0.481$\pm$0.130 & 0.123$\pm$0.092 & 0.559 & 0.337 \\
    All Ref      & 0.508$\pm$0.110 & 0.128$\pm$0.121 & 0.642 & 0.415 \\
    \bottomrule
  \end{tabular}
\end{table}

\subsection{Candidate Dataset Proxy Labels}
\label{sec:appendix_full_labels}

Full candidate dataset proxy labels are listed in Table~\ref{tab:full_candidate_dataset_labels}. 
To reduce space, datasets with identical labels and contiguous naming patterns are merged into grouped entries. 
In grouped rows, the reported labels apply to all datasets in the indicated range. 
Here, ``Y'' denotes documented exposure in official sources, and ``N'' denotes no documented evidence in official sources.
\begingroup
\scriptsize
\setlength{\tabcolsep}{2.8pt}
\renewcommand{\arraystretch}{0.95}
\begin{longtable}{p{3.1cm}cccccc p{3.1cm}cccccc}
\caption{Full candidate--dataset proxy labels for the evaluated benchmark. Column abbreviations: Ch = Chronos, Ka = Kairos, M1 = Moirai1, M2 = Moirai2, TF = TimesFM2.0, Ti = TiRex.}
\label{tab:full_candidate_dataset_labels}\\
\toprule
\textbf{Dataset} & \textbf{Ch} & \textbf{Ka} & \textbf{M1} & \textbf{M2} & \textbf{TF} & \textbf{Ti} &
\textbf{Dataset} & \textbf{Ch} & \textbf{Ka} & \textbf{M1} & \textbf{M2} & \textbf{TF} & \textbf{Ti} \\
\midrule
\endfirsthead

\toprule
\textbf{Dataset} & \textbf{Ch} & \textbf{Ka} & \textbf{M1} & \textbf{M2} & \textbf{TF} & \textbf{Ti} &
\textbf{Dataset} & \textbf{Ch} & \textbf{Ka} & \textbf{M1} & \textbf{M2} & \textbf{TF} & \textbf{Ti} \\
\midrule
\endhead

\bottomrule
\endfoot

alibaba\_cluster\_trace\_2018 & N & Y & Y & Y & N & N &
azure\_vm\_traces\_2017 & N & Y & Y & Y & N & N \\

bdg-2\_bear & N & Y & Y & Y & N & N &
bdg-2\_fox & N & Y & Y & Y & N & N \\

bdg-2\_panther & N & Y & Y & Y & N & N &
bdg-2\_rat & N & Y & Y & Y & N & N \\

BEIJING\_SUBWAY\_30MIN & N & Y & Y & Y & N & N &
bitbrains\_fast\_storage/H & N & N & N & N & N & N \\

bitbrains\_rnd/H & N & N & N & N & N & N &
borg\_cluster\_data\_2011 & N & Y & Y & Y & N & N \\

buildings\_900k & N & Y & Y & Y & N & N &
bull & N & Y & Y & Y & N & N \\

cdc\_fluview\_ilinet & N & Y & Y & Y & N & N &
cdc\_fluview\_who\_nrevss & N & Y & Y & Y & N & N \\

china\_air\_quality & N & Y & Y & Y & N & N &
cmip6\_1850--2010 & N & Y & Y & Y & N & N \\

covid\_deaths & Y & N & Y & N & N & Y &
covid\_mobility & N & Y & Y & Y & N & N \\

electricity/H & Y & N & N & N & Y & Y &
electricity/W & Y & N & N & N & Y & Y \\

era5\_1989--2018 & N & Y & Y & Y & N & N &
extended\_web\_traffic\_with\_missing & N & Y & Y & Y & Y & N \\

favorita\_sales & N & Y & Y & Y & Y & N &
favorita\_transactions & N & Y & Y & Y & N & N \\

fred\_md & Y & Y & Y & Y & N & Y &
hierarchical\_sales/D & N & N & Y & N & N & N \\

HZMETRO & N & Y & Y & Y & N & N &
ideal & N & Y & Y & Y & N & N \\

kdd\_cup\_2018\_with\_missing/D & N & N & Y & N & N & N &
kdd\_cup\_2018\_with\_missing/H & N & N & Y & N & N & N \\

kdd2022 & N & Y & Y & Y & N & N &
largest\_2017--2021 & N & Y & Y & Y & N & N \\

lcl & N & Y & Y & Y & N & N &
london\_smart\_meters\_with\_missing & N & Y & Y & Y & N & N \\

LOOP\_SEATTLE/H & N & N & Y & N & N & N &
LOS\_LOOP & N & Y & Y & Y & N & N \\

M\_DENSE/H & N & N & Y & N & N & N &
m4\_daily & Y & N & Y & N & Y & Y \\

m4\_hourly & Y & N & Y & N & Y & Y &
m4\_monthly & Y & N & Y & N & Y & Y \\

m4\_quarterly & Y & N & Y & N & Y & Y &
m4\_weekly & Y & N & Y & N & N & Y \\

m4\_yearly & Y & N & Y & N & Y & Y &
m5 & Y & Y & Y & Y & N & Y \\

nn5\_daily\_with\_missing & Y & Y & Y & Y & N & Y &
pedestrian\_counts & Y & Y & Y & Y & N & Y \\

PEMS\_BAY & N & Y & Y & Y & N & N &
PEMS03 & N & Y & Y & Y & N & N \\

PEMS04 & N & Y & Y & Y & N & N &
PEMS07 & N & Y & Y & Y & N & N \\

PEMS08 & N & Y & Y & Y & N & N &
project\_tycho & N & Y & Y & Y & N & N \\

Q-TRAFFIC & N & Y & Y & Y & Y & N &
residential\_load\_power & N & Y & Y & Y & N & N \\

residential\_pv\_power & N & Y & Y & Y & N & N &
restaurant & N & N & Y & N & N & N \\

rideshare\_with\_missing & N & Y & Y & Y & N & N &
SHMETRO & N & Y & Y & Y & N & N \\

solar/D & Y & N & N & N & N & Y &
subseasonal & N & Y & Y & Y & N & N \\

subseasonal\_precip & N & Y & Y & Y & N & N &
SZ\_TAXI/H & N & N & Y & N & N & N \\

taxi\_30min & Y & Y & Y & Y & N & Y &
temperature\_rain\_with\_missing & N & N & Y & N & N & N \\

tourism\_monthly & Y & Y & Y & Y & N & Y &
traffic\_hourly & N & Y & Y & Y & Y & N \\

uber\_tlc\_daily & Y & Y & Y & Y & N & Y &
uber\_tlc\_hourly & Y & Y & Y & Y & N & Y \\

vehicle\_trips\_with\_missing & N & Y & Y & Y & N & N &
weather & Y & Y & Y & Y & Y & Y \\

wiki-rolling\_nips & N & Y & Y & Y & N & N &
wind\_farms\_with\_missing & N & Y & Y & Y & N & N \\
\end{longtable}
\endgroup

\subsection{Family Grouping and Aggregation Levels}
\label{sec:appendix_grouping}

For data splitting, we retain family metadata that records which datasets belong to the same benchmark family. Under \texttt{group\_split=family}, datasets from the same family are assigned to the same split. Under \texttt{group\_split=none}, this grouping constraint is removed. We use the family split by default because it keeps closely related datasets on the same side of the split and reduces cross-split transfer through renamed or lightly transformed variants.

We report both \emph{Pair-MICRO} and \emph{Model-MACRO} summaries. Pair-MICRO pools all candidate--dataset test decisions and then computes the evaluation metric on the pooled set. Model-MACRO first computes the metric separately for each candidate TSFM and then averages across candidates. Pair-MICRO reflects pooled behavior over all audit pairs, whereas Model-MACRO avoids overweighting candidates with larger numbers of auditable datasets.

\section{Experimental Protocol and Implementation Details}
\label{sec:appendix_protocol_details}

\subsection{Audit Protocol and \texorpdfstring{FP-$k$}{FP-k} Calibration}
\label{sec:appendix_protocol}

This subsection complements Sections~\ref{sec:method} and~\ref{sec:calibration_decision} by specifying how the audit protocol is instantiated in the evaluation code. In particular, it records the repeated split construction, calibration-set composition, epoch selection rule, and the operational form of FP-$k$ threshold calibration used throughout the appendix.

\paragraph{Repeated calibration/test splits.}
The atomic label-bearing example is the candidate--dataset pair $(f_{\bm{\theta}}, \mathcal{D})$. Evaluation is performed over repeated calibration/test splits of these audit pairs. In the default setting, we use 30 repeated splits, random seed 0, and \texttt{group\_split=family}. For repeat index $r$, the random generator is initialized with seed $0 + 10007\,r$. Split generation is retried until both proxy-label classes appear in the calibration split.

\paragraph{Calibration-set composition.}
The calibration split is constructed to prioritize clean audit pairs. By default, it targets approximately 80\% of the clean calibration pairs and then samples positive pairs at a 1:1 ratio relative to the number of clean calibration pairs. This design reflects the conservative role of calibration under the FP-$k$ protocol, where the threshold is determined from clean calibration scores.

\paragraph{Shared split manifests across methods.}
For a fixed candidate and protocol setting, all baselines and all \textsc{TSFMAudit} variants are evaluated on the same repeated split manifests. This includes the candidate-only baseline, the single-reference variants, and the all-reference variants. As a result, appendix comparisons isolate differences in the scoring rule rather than differences in split construction or threshold calibration.

\paragraph{Epoch selection within each repeat.}
For each repeated split, we evaluate the epochwise scorers $\{g_t\}_{t=1}^{T_{\mathrm{probe}}}$ and select the epoch with the highest calibration MCC, following Section~\ref{sec:calibration_decision}. Unless otherwise stated, this selection is performed separately within each repeat. The default epochwise scorer is logistic regression.

\paragraph{FP-$k$ threshold calibration.}
Let $S(f_{\bm{\theta}},\mathcal{D})$ denote the selected continuous score for an audit pair under the chosen epoch. The threshold is determined only from clean calibration pairs,
\[
\mathcal{V}_{\mathrm{clean}}
=
\left\{
(f_{\bm{\theta}},\mathcal{D}) :
\widetilde{C}(f_{\bm{\theta}},\mathcal{D}) = 0
\right\}.
\]
Under FP-0, the threshold is set to the largest clean calibration score. Under FP-1, it is set to the second-largest clean calibration score. More generally, FP-$k$ uses the $(k+1)$-st largest clean calibration score as the threshold. The final binary decision is
\[
\widehat{S}(f_{\bm{\theta}},\mathcal{D})
=
\mathbb{I}\!\left[
S(f_{\bm{\theta}},\mathcal{D}) > \tau_k
\right].
\]
The main paper uses FP-0 unless otherwise stated.

\subsection{Reference Suite and Baseline Scope}
\label{sec:appendix_baselines}

The six audited candidates are Chronos, TiRex, Moirai1, Moirai2, TimesFM2.0, and Kairos. The reference suite used throughout the main experiments is
\[
\mathcal{R}=\{\text{ScratchCNN},\text{ScratchTransformer},\text{Stat},\text{VisionTS}\}.
\]
Appendix comparisons include the candidate-only baseline, the single-reference variants, and the all-reference variants defined from this reference suite. We use the same all-reference abbreviations as Section~\ref{sec:setup}: All Ref concatenates the difference and ratio features over all references, while All Ref+Det additionally appends the direct candidate dynamics (Det), i.e., the candidate trace summary.

\paragraph{Default scorer implementation.}
Unless otherwise stated, the default epochwise scorer is logistic regression with a standard-scaling front end, \texttt{liblinear} solver, balanced class weights, and \texttt{max\_iter=1000}. In addition to this default scorer, the appendix reports detector-family ablations with random forests~\cite{Breiman2001RandomForests} and gradient-boosted trees. These ablations are included to assess detector sensitivity and are not part of the default protocol.

\paragraph{Evaluation metrics.}
The evaluation code reports AUROC, AUPRC, Macro-F1, MCC, and balanced accuracy. The main paper emphasizes MCC, Macro-F1, AUROC, and balanced accuracy. We additionally report AUPRC in the appendix because the class balance is often highly skewed.

\paragraph{Frequency-domain baseline.}
\label{sec:appendix_fft_baseline}
We additionally report a frequency-domain heuristic baseline, denoted \texttt{ts\_mink\_fft\_*}. 
Given residuals $r_t = y_t - \hat{y}_t$, we compute the frequency energy spectrum $|\mathrm{FFT}(r)|^2$ and define a series-level score as the mean energy of the top-$K\%$ frequency components, with $K\in\{10,20,30\}$. 
The dataset-level score is obtained by averaging these series-level scores over the audited dataset. 
This baseline is included as a time-series analogue of Min-K\% style~\cite{shi2024detecting} tests and is evaluated under the same six-candidate protocol as the main comparison.

\subsection{Probe Hyperparameters and Trace Construction}
\label{sec:appendix_probe}

This subsection records the shared probe settings and the implementation choices used to construct the trace summaries in Section~\ref{sec:method}. Table~\ref{tab:repro_details} summarizes the protocol and implementation choices that are fixed throughout the experiments.

\paragraph{Shared defaults.}
Unless otherwise stated, we use $T_{\mathrm{probe}}=10$ probe epochs, batch size 4, and the context--horizon pair $(L,H)=(128,24)$ following GIFT-Eval. The default learning rate is $10^{-3}$, while TimesFM2.0 uses $10^{-4}$ for stability.

\paragraph{Window-level preprocessing.}
Each context--target window $(\mathbf{H}_i,\mathbf{Y}_i)$ is normalized using the mean and standard deviation of the context segment $\mathbf{H}_i$. Concretely, the context is standardized by its own statistics, and the target is normalized using the same context statistics. This keeps the probe dynamics comparable across datasets with substantially different scales.

\paragraph{Optimization details.}
The probing procedure uses AdamW~\cite{loshchilov2019adamw}, mean squared forecasting loss, weight decay $0.01$, and gradient clipping with maximum norm 1.0. The initial loss $\ell_0$ is recorded before any parameter updates, and subsequent losses are recorded after each probe epoch. The parameter displacement $w_t$ is computed as the $\ell_2$ norm between the current parameter state and the checkpoint state at probe start. Before auditing a new dataset, the model is reset to the same initial checkpoint so that all candidate--dataset traces are measured from a common origin.

\paragraph{Trace summaries.}
At each probe epoch $t$, the implementation records the loss $\ell_t$ and parameter displacement $w_t$, and then computes the relative loss drop
\[
d_t = \frac{\ell_0-\ell_t}{\ell_0+\varepsilon_\ell},
\]
and the adaptation efficiency
\[
a_t = \frac{d_t}{w_t+\varepsilon_{\mathrm{ae}}}.
\]
These quantities define the candidate and reference trace summaries used in the main method,
\[
\mathbf{x}(\mathcal{D}) = [w_t,d_t,a_t]_{1\le t\le T_{\mathrm{probe}}},
\qquad
\mathbf{x}^{(r)}(\mathcal{D}) = [w_t^{(r)},d_t^{(r)},a_t^{(r)}]_{1\le t\le T_{\mathrm{probe}}}.
\]
The reference-based features are then constructed from the per-epoch differences and ratios defined in Section~\ref{sec:relative_debiasing}. Additional recorded quantities such as gradient norm are used only in auxiliary implementation diagnostics and are not part of the default trace representation.

\begin{table}[t]
\centering
\small
\begin{tabular}{ll}
\toprule
\textbf{Item} & \textbf{Value} \\
\midrule
Audit unit & candidate--dataset pair $(f_{\bm{\theta}},\mathcal{D})$ \\
Split repeats & 30 \\
Default split seed & 0 \\
Family-aware split & \texttt{group\_split=family} \\
Calibration clean fraction & 0.8 \\
Calibration leaked:clean ratio & 1.0 \\
Main threshold rule & FP-0 \\
Default epochwise scorer & Logistic regression + standard scaling \\
Logistic-regression settings & \texttt{liblinear}, balanced class weights, \texttt{max\_iter=1000} \\
Probe length & $T_{\mathrm{probe}}=10$ \\
Batch size & 4 \\
Default context / horizon & $(128,24)$ \\
Window normalization & context-wise z-score \\
Probe optimizer & AdamW \\
Weight decay & 0.01 \\
Gradient clipping & max norm 1.0 \\
Per-dataset model reset & yes \\
Efficiency stabilizer $\varepsilon_{\mathrm{ae}}$ & $10^{-6}$ \\
Reference ratio stabilizer $\varepsilon_{\mathrm{ref}}$ & $10^{-6}$ \\
\bottomrule
\end{tabular}
\caption{Protocol and implementation details fixed throughout the experiments.}
\label{tab:repro_details}
\end{table}

\section{Additional Main Results}
\label{sec:appendix_additional_results}

\subsection{Sensitivity to FP Budget and Group Splits}
\label{sec:appendix_sensitivity}

\begin{figure*}[t]
  \centering
  \includegraphics[width=\linewidth]{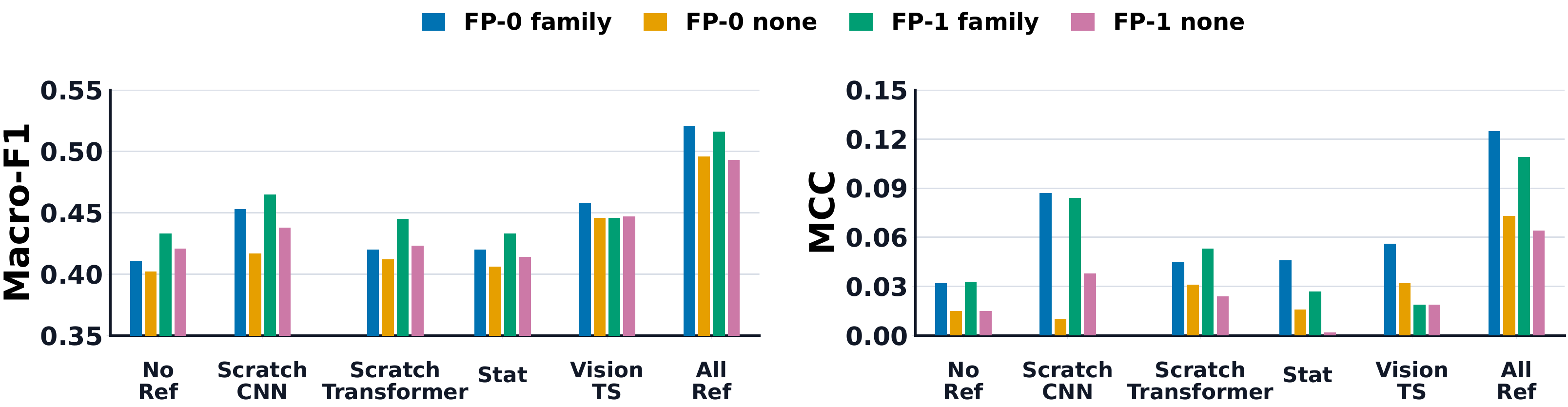}
  \caption{Sensitivity to the false-positive budget and split grouping on Model-MACRO aggregation. Bars report mean Macro-F1 and MCC over repeated splits. FP-0 uses the most conservative clean-score threshold, while FP-1 allows one clean calibration exceedance. Family grouping is slightly more stringent than dataset-level splitting, but the advantage of reference-based debiasing remains stable across all protocol variants.}
  \label{fig:sensitivity_fpksplit_macro}
\end{figure*}

\begin{figure*}[t]
  \centering
  \includegraphics[width=\linewidth]{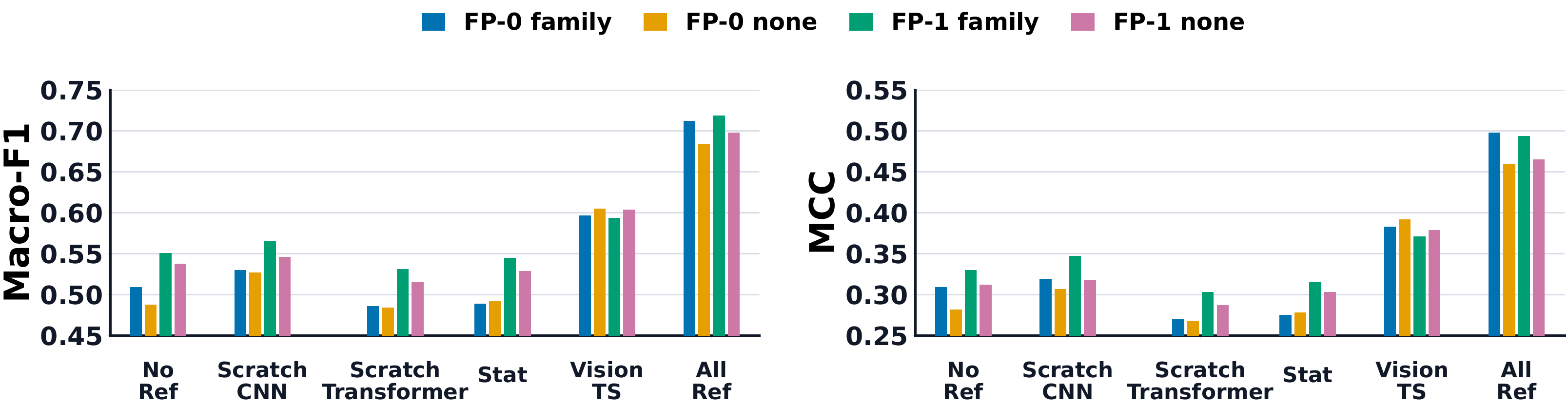}
  \caption{Sensitivity to the false-positive budget and split grouping on Pair-MICRO aggregation. Bars report Macro-F1 and MCC after pooling all candidate--dataset pairs. FP-0 uses the most conservative clean-score threshold, while FP-1 allows one clean calibration exceedance. Across all protocol variants, reference-based methods consistently outperform the candidate-only baseline, with VisionTS and All Ref remaining particularly strong.}
  \label{fig:sensitivity_fpksplit_micro}
\end{figure*}

Figures~\ref{fig:sensitivity_fpksplit_macro} and~\ref{fig:sensitivity_fpksplit_micro} examine sensitivity to two protocol choices: the false-positive budget, comparing FP-0 and FP-1, and the split grouping strategy, comparing \texttt{group\_split=family} and \texttt{group\_split=none}. The two figures report the same comparison under Model-MACRO and Pair-MICRO aggregation, respectively.


\paragraph{Findings.}
The main qualitative conclusions remain stable under both FP budgets and both grouping rules. Reference-based variants consistently improve over the candidate-only baseline under both FP-0 and FP-1. The strongest single-reference choice depends on the aggregation level: VisionTS remains particularly strong on Pair-MICRO, while ScratchCNN remains competitive on Model-MACRO MCC under family-aware splits. Although \texttt{group\_split=family} is slightly more stringent than \texttt{group\_split=none}, it better matches the intended unknown-dataset setting because related benchmarks are kept in the same split. Figures~\ref{fig:sensitivity_fpksplit_macro} and~\ref{fig:sensitivity_fpksplit_micro} plot Macro-F1 and MCC separately for readability.

\subsection{Loss-Based Restricted Variant}
\label{sec:appendix_lossonly}

We additionally evaluate a restricted variant of \textsc{TSFMAudit} that uses only loss-based trace quantities. 
This variant retains features derived from the loss trajectory and the relative loss drop, and removes parameter-displacement quantities and adaptation-efficiency quantities. 
Table~\ref{tab:lossonly_ablation} compares this restricted variant against the default full trace representation under the same FP-0 protocol and family-aware splits used in the main appendix tables.

\begin{table*}[t]
  \caption{Loss-based restricted variant under FP-0 auditing with family-aware splits. Each cell reports Macro-F1 / MCC. The restricted variant remains informative, but the full trace representation is generally stronger, with the largest gap appearing for VisionTS on Pair-MICRO.}
  \label{tab:lossonly_ablation}
  \centering
  \small
  \setlength{\tabcolsep}{4pt}
  \begin{tabular}{@{}lcccc@{}}
    \toprule
    \textbf{Reference} & \textbf{MACRO (full)} & \textbf{MACRO (loss-based)} & \textbf{MICRO (full)} & \textbf{MICRO (loss-based)} \\
    \midrule
    No ref     & 0.411 / 0.032 & 0.371 / 0.003  & 0.509 / 0.309 & 0.432 / 0.242 \\
    ScratchCNN & 0.453 / 0.087 & 0.436 / 0.108  & 0.530 / 0.319 & 0.469 / 0.275 \\
    ScratchTransformer & 0.420 / 0.045 & 0.399 / 0.045  & 0.486 / 0.270 & 0.446 / 0.245 \\
    Stat       & 0.420 / 0.046 & 0.383 / 0.015  & 0.489 / 0.275 & 0.442 / 0.248 \\
    VisionTS   & 0.458 / 0.056 & 0.380 / -0.006 & 0.597 / 0.383 & 0.452 / 0.247 \\
    All Ref    & 0.521 / 0.125 & 0.515 / 0.117  & 0.712 / 0.498 & 0.683 / 0.372 \\
    \bottomrule
  \end{tabular}
\end{table*}
\paragraph{Findings.}
The loss-based restricted variant remains informative, but it generally underperforms the full trace representation. 
The size of the gap depends on the reference model. 
With ScratchCNN as reference, the restricted variant slightly improves Model-MACRO MCC from $0.087$ to $0.108$, while the corresponding Pair-MICRO MCC decreases from $0.319$ to $0.275$. 
By contrast, the gap is larger for VisionTS, where Pair-MICRO MCC drops from $0.383$ to $0.247$. 
Taken together, these results suggest that loss-based quantities alone retain useful audit signal, but the default full representation is more reliable overall

\subsection{Per-Candidate Reference Effects}
\label{sec:appendix_per_candidate}

\begin{figure}[t]
  \centering
  \includegraphics[width=\linewidth]{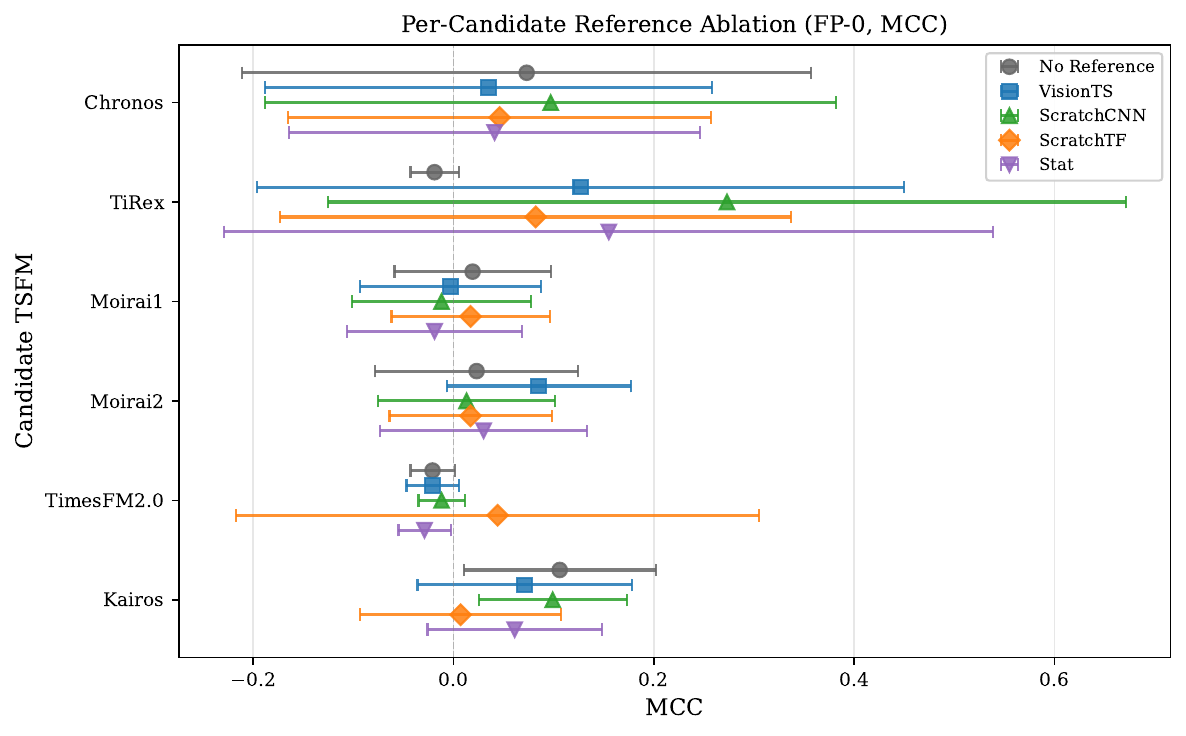}
  \caption{Per-candidate reference ablation under FP-0 calibration with family-aware splits. Reference utility is candidate dependent: ScratchCNN yields the clearest gain for TiRex, VisionTS is strongest for Moirai2, while several candidates show only modest or inconsistent changes across references.}
  \label{fig:per_candidate_ablation}
\end{figure}

Figure~\ref{fig:per_candidate_ablation} reports candidate-specific MCC under FP-0 calibration with family-aware splits. The aggregate results reported elsewhere in the appendix average over substantial heterogeneity across candidates, and this figure makes that heterogeneity explicit.

\paragraph{Findings.}
The benefit of reference-based debiasing is not uniform across candidates. TiRex shows the largest gain, with ScratchCNN increasing MCC from approximately $-0.02$ under the no-reference baseline to about $0.27$. Moirai2 benefits most from VisionTS, which improves over the no-reference baseline by a smaller but consistent margin. By contrast, Chronos and Moirai1 show only modest differences across references, and TimesFM2.0 does not exhibit a stable average gain from any single reference. Kairos also does not benefit on average from the current references, with the clearest degradation appearing under ScratchTransformer relative to the no-reference baseline. Taken together, these patterns indicate that reference quality is candidate dependent: a useful reference is one whose probe-time behavior provides a plausible counterfactual for the candidate under audit.

\paragraph{Interpretation under label imbalance.}
This candidate-level heterogeneity should be interpreted together with proxy-label balance. Under strict FP-0 calibration, threshold selection depends only on clean calibration pairs, so candidates with very few documented-clean audit pairs can show higher across-split variance. This is particularly relevant for Moirai1 under the current proxy labels. For this reason, the candidate-level patterns in Figure~\ref{fig:per_candidate_ablation} should be read together with the FP-0 and FP-1 sensitivity results in Figures~\ref{fig:sensitivity_fpksplit_macro} and~\ref{fig:sensitivity_fpksplit_micro}.

\subsection{Probe Length Sensitivity}
\label{sec:appendix_probe_length}

\begin{figure*}[t]
  \centering
  \includegraphics[width=\linewidth]{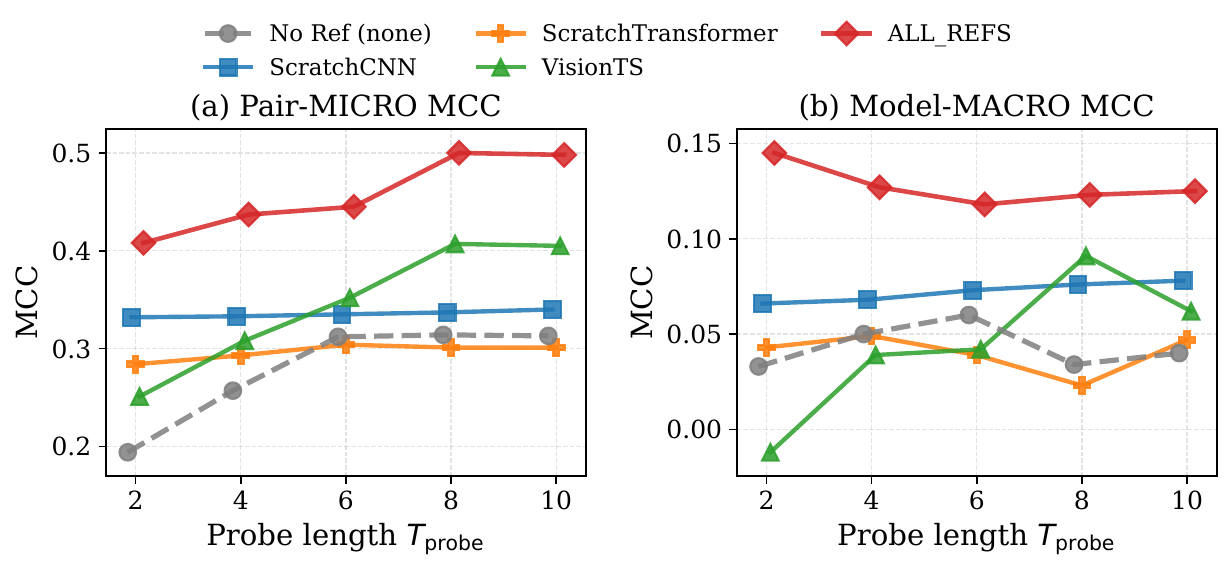}
  \caption{Probe length sensitivity under FP-0 calibration with family-aware splits. The combined figure reports Model-MACRO MCC and Pair-MICRO MCC as functions of $T_{\mathrm{probe}}$. Useful audit signal appears early, while most configurations change only modestly beyond moderate probe lengths.}
  \label{fig:probe_sensitivity_combined_mcc}
\end{figure*}

Figure~\ref{fig:probe_sensitivity_combined_mcc} reports Model-MACRO MCC and Pair-MICRO MCC as functions of probe length $T_{\mathrm{probe}} \in \{2,4,6,8,10\}$ under FP-0 calibration with family-aware splits. The goal is to determine whether useful audit signal already appears under short probing, and whether longer probes materially change the relative ranking of representative reference configurations.

\paragraph{Findings.}
Useful audit signal appears early, but very short probes can leave part of that signal unrecovered. In Figure~\ref{fig:probe_sensitivity_combined_mcc}, the candidate-only baseline increases from approximately $0.19$ at $T_{\mathrm{probe}}=2$ to about $0.31$ at $T_{\mathrm{probe}}=6$, and then remains essentially unchanged through $T_{\mathrm{probe}}=8$ and $10$ in Pair-MICRO MCC. A similar stabilization is visible for ScratchCNN, whose Pair-MICRO MCC changes only slightly across the full range.

The configurations that benefit most from longer probes are VisionTS and All Ref. VisionTS improves steadily in Pair-MICRO MCC from about $0.25$ at $T_{\mathrm{probe}}=2$ to about $0.41$ at $T_{\mathrm{probe}}=8$, after which it is effectively flat. The All Ref variant shows the same pattern, increasing from about $0.41$ at $T_{\mathrm{probe}}=2$ to about $0.50$ at $T_{\mathrm{probe}}=8$, with no material gain at $T_{\mathrm{probe}}=10$. This indicates that the main pooled detection gains are usually realized by moderate probe lengths rather than by the longest probe considered here.

The same figure shows a similar but less monotone pattern at the Model-MACRO level. ScratchCNN is comparatively stable across probe lengths, while VisionTS and Stat improve more substantially from short to moderate probes. At the same time, the All Ref variant remains the strongest overall configuration across all tested probe lengths, even though its Model-MACRO MCC varies only within a relatively narrow range. Taken together, these results suggest that $T_{\mathrm{probe}}=10$ is a stable default that lies near the performance plateau, rather than a finely tuned optimum.

\subsection{TIME Benchmark Practical Evaluation Details}
\label{sec:appendix_time}

This subsection records the implementation details of the deployment-oriented evaluation on the TIME Benchmark~\cite{qiao2026time} reported in Section~\ref{sec:time_negative_control}. The guiding principle is that no part of the audit pipeline is re-fit to TIME: the selected scorer and threshold are inherited from the main experiment, and TIME serves only as the deployment target.

\paragraph{Scorer and threshold reuse.}
We adopt the All Ref configuration, which is the strongest aggregate setting in Table~\ref{tab:main_results}. The selected scorer $g^{\ast}$ is the epochwise logistic-regression scorer chosen by the calibration-MCC rule of Section~\ref{sec:calibration_decision}, with the standard-scaling front end and solver settings recorded in Appendix~\ref{sec:appendix_baselines}. The threshold $\tau_0$ is its FP-$0$ threshold, obtained on the main-experiment validation set from the clean calibration scores under the protocol of Appendix~\ref{sec:appendix_protocol}. Before applying $\left(g^{\ast},\tau_0\right)$ to TIME, we verify on the same validation set that this fixed pair reproduces the All Ref behavior reported in Table~\ref{tab:main_results}. The same pair is then applied to TIME without any further fitting; in particular, no TIME data enters scorer training or threshold calibration.

\paragraph{Probing on TIME.}
For each candidate TSFM $f_{\bm{\theta}}$ and each TIME dataset $\mathcal{D}$, we run the same fine tuning protocol used in the main experiment. Concretely, we use $T_{\mathrm{probe}}=10$ probe epochs, batch size $4$, context--horizon pair $(L,H)=(128,24)$, AdamW with the candidate-specific learning rate listed in Appendix~\ref{sec:appendix_probe}, context-wise z-score normalization of each window, and per-dataset model reset to the same initial checkpoint used in the main experiment. Each reference model $f_{\bm{\theta}^{(r)}}\in\mathcal{R}=\{\text{ScratchCNN},\text{ScratchTransformer},\text{Stat},\text{VisionTS}\}$ is probed on TIME under identical settings. Trace summaries $\mathbf{x}(\mathcal{D})$ and $\mathbf{x}^{(r)}(\mathcal{D})$ and the debiased features $\{\bm{\Delta}_t^{(r)}(\mathcal{D}),\bm{\rho}_t^{(r)}(\mathcal{D})\}_{r=1}^{|\mathcal{R}|}$ are constructed exactly as in Section~\ref{sec:relative_debiasing}. The previously calibrated $\left(g^{\ast},\tau_0\right)$ then produces a binary audit decision $\widehat{S}\!\left(f_{\bm{\theta}},\mathcal{D}\right)$ for each audit pair.

\paragraph{Evaluation set and confidence bounds.}
Each of the six candidates is audited on $50$ TIME datasets, yielding $300$ audit pairs $\left(f_{\bm{\theta}},\mathcal{D}\right)$ in total. The per-candidate rates and one-sided $95\%$ Clopper--Pearson lower bounds in Table~\ref{tab:time_negative_control} are computed under the standard binomial model on the $50$ decisions per candidate; the pooled lower bound is computed analogously over the $300$ pooled decisions. As noted in Section~\ref{sec:time_negative_control}, the independence assumption underlying these bounds holds only approximately because the $50$ TIME datasets share a common evaluation protocol, so the bounds should be read as descriptive rather than as a formal coverage guarantee.

\end{document}